\documentclass[runningheads]{llncs}
\usepackage{graphicx}

\usepackage{tikz}
\usepackage{comment}
\usepackage{amsmath,amssymb} %
\usepackage{color}
\usepackage{adjustbox}

\usepackage[accsupp]{axessibility}  %

\usepackage{amssymb}             %
\usepackage{amsfonts}            %
\usepackage{mathrsfs}            %
\usepackage{url}
\usepackage[linesnumbered,ruled,vlined]{algorithm2e}

\usepackage{booktabs}
\usepackage{multirow}

\usepackage{xcolor}
\usepackage[normalem]{ulem}

\newcommand\delete{\bgroup\markoverwith{\textcolor{red}{\rule[0.5ex]{2pt}{1.0pt}}}\ULon}

\usepackage{pifont}
\newcommand{\cmark}{\ding{51}}%

\usepackage[pagebackref,breaklinks,colorlinks]{hyperref}
\usepackage[capitalize]{cleveref}
\usepackage{rotating}

\DeclareMathOperator*{\argmin}{arg\,min}

\begin{document}
\pagestyle{headings}
\mainmatter
\def\ECCVSubNumber{2609}  %

\title{Learning Semantic Correspondence with Sparse Annotations} %

\titlerunning{Learning Semantic Correspondence with Sparse Annotations}

\author{Shuaiyi Huang\inst{1} \and
Luyu Yang\inst{1} \and
Bo He\inst{1} \and
Songyang Zhang\inst{2}\and \\
Xuming He\inst{3,4} \and
Abhinav Shrivastava\inst{1}} 

\authorrunning{S. Huang et al.}
\institute{University of Maryland, College Park \and Shanghai AI Laboratory \and ShanghaiTech University \and Shanghai Engineering Research Center of Intelligent Vision and Imaging
\\
\email{\{huangshy,loyo,bohe\}@umd.edu, zhangsongyang@pjlab.org.cn, hexm@shanghaitech.edu.cn, abhinav@cs.umd.edu}
}
\maketitle
\begin{abstract}
Finding dense semantic correspondence is a fundamental problem in computer vision, which remains challenging in complex scenes due to background clutter, extreme intra-class variation, and a severe lack of ground truth. In this paper, we aim to address the challenge of label sparsity in semantic correspondence by enriching supervision signals from sparse keypoint annotations. 
To this end, we first propose a teacher-student learning paradigm for generating dense pseudo-labels and then develop two novel strategies for denoising pseudo-labels. In particular, we use spatial priors around the sparse annotations to suppress the noisy pseudo-labels. In addition, we introduce a loss-driven dynamic label selection strategy for label denoising. We instantiate our paradigm with two variants of learning strategies: a single offline teacher setting, and mutual online teachers setting. Our approach achieves notable improvements on three challenging benchmarks for semantic correspondence and establishes the new state-of-the-art. Project page:~\url{https://shuaiyihuang.github.io/publications/SCorrSAN}. 
\end{abstract}

\section{Introduction}

Estimating pixel-wise correspondence between images is a fundamental task in computer vision applications. Correspondences like stereo disparities~\cite{scharstein2002taxonomy} and optical flow~\cite{horn1981determining} are widely used for applications such as surface reconstruction and video analysis~\cite{chauhan2013moving,goldstein2012video}. Recently, such instance-level dense correspondence has been generalized to semantic correspondence, which, given a pair of images, aligns the object instance from the first image to the one of the same category in the second image~\cite{Rocco2017,Rocco2018,kim2017fcss,huang2019dynamic,min2019hyperpixel,min2020learning,jeon2018parn,truong2020glu}. It has attracted growing attention due to its practical use in segmentation, style-transfer, and image editing~\cite{lan2021discobox,chen2020show,lee2020reference,dale2009image,jeon2019joint,hong2022cost}. However, background clutter, intra-class variations, viewpoint changes, and particularly the severe lack of annotations make it an extremely challenging task.

\begin{figure}[t]
    \centering
    \adjincludegraphics[width=\linewidth, trim={{0\width} {0.02\height} {0\width} {0\height}},clip]{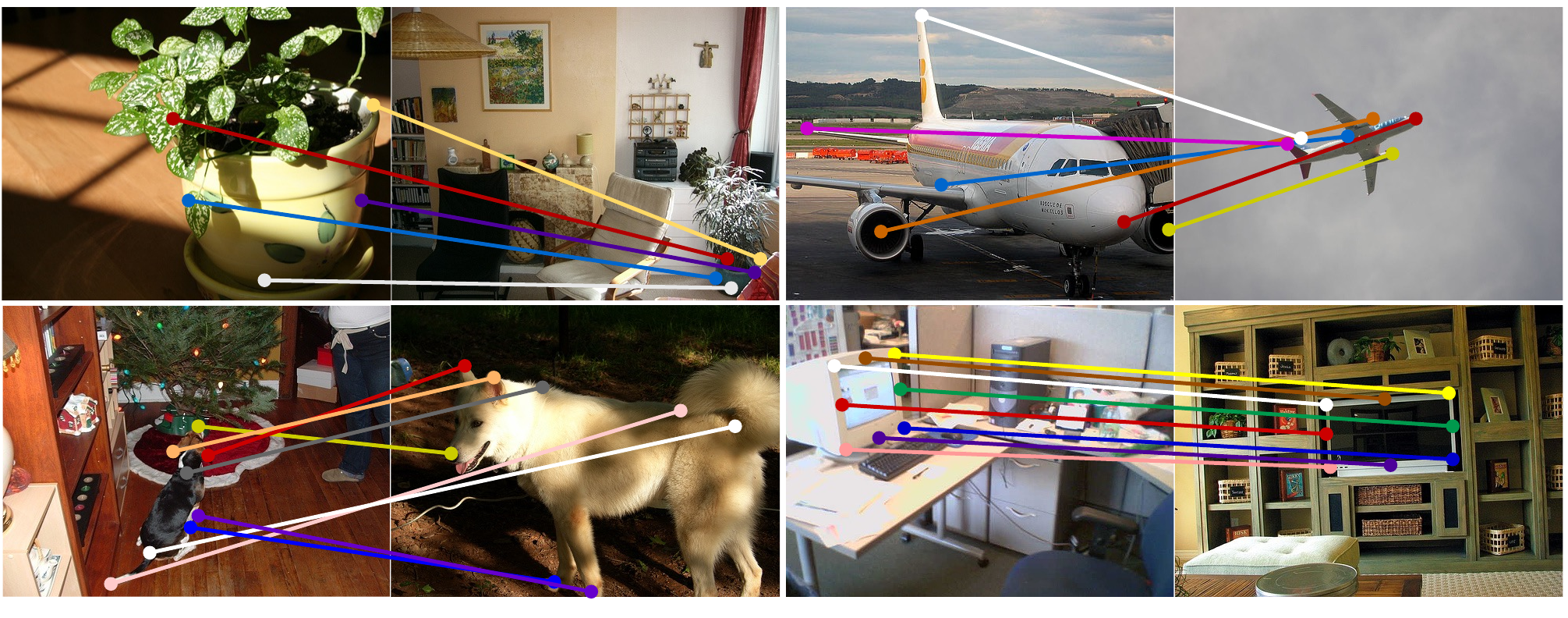}
    \caption{\textbf{Motivation.} Image pairs from SPair-71k dataset~\cite{min2019spair} training split show sparse annotations for semantic correspondence.}
    \label{fig:ad}
\end{figure}

Due to the high cost of dense annotation, the semantic correspondence task only provides sparse keypoint annotations in the supervised setting~\cite{min2020learning,li2020correspondence,li2021probabilistic,lee2021patchmatch} as shown in Fig.~\ref{fig:ad}. In this paper, we are motivated by how to better utilize the limited supervision. Specifically, we explore the techniques to generate pseudo-labels. However, due to the inevitably noisy effect of pseudo-labels, filtering out noisy pseudo-labels remains a challenging problem. Our key observation is that sparse keypoint annotations and their neighborhood encode rich semantic information. By utilizing this spatial prior, one can seek reliable pseudo-labels that are more likely in the foreground region of interest.

To this end, we propose a novel teacher-student framework to cope with label sparsity. The teacher model is trained with sparse keypoint annotations to generate dense pseudo-labels. To improve pseudo-labels quality, we propose (a) using the sparse annotations as spatial prior to suppress the noisy pseudo-labels, and (b) loss-driven dynamic label selection. To train the models, we propose two variants of our strategy: (1) a single offline teacher with an online student, and (2) two online teachers that learn from each other. Both variants lead to substantial performance improvements over the state-of-the-art.

We instantiate our novel learning strategy based on our proposed simple, yet effective network architecture for semantic correspondence. The proposed network comprises three modules: (a) a feature extractor equipped with our efficient spatial context encoder, (b) a parameter-free correlation map module, and (c) a flow estimator with our designed high-resolution loss.

The contributions are summarized as follows:
\begin{itemize}
    \item We propose a simple, yet effective model for semantic correspondence without any transformer or 4D-conv for correlation refinement. The key ingredients are an efficient spatial context encoder and a high-resolution loss.
    \item We introduce a novel teacher-student learning paradigm to enrich the supervision guidance when only sparse annotations are available. Two key techniques are a novel spatial-prior based label filtering and a loss-driven dynamic label selection strategy for high-quality pseudo-label generation.
    \item Our novel learning strategy is simple to implement, and achieves state-of-the-art results with good generalization performance on three semantic correspondence benchmarks, demonstrating the effectiveness of our method. 
\end{itemize}

\section{Related Work}

\subsection{Semantic Correspondence}
Conventional approaches for semantic correspondence mostly employ hand-crafted features together with geometric models~\cite{liu2011sift,tola2010daisy,taniai2016joint}. These methods establish correspondences across images via energy minimization. SIFT Flow~\cite{liu2011sift} pioneers the idea of finding correspondences across similar scenes with SIFT descriptors. Ham~\textit{et al.}~\cite{ham2018proposal} utilize object proposals as the matching primitives and establish correspondence via HOG descriptors. Those methods often have difficulty dealing with background clutter, intra-class variations, and large viewpoint changes due to the lack of semantics in features.

Recently, deep CNN-based methods have been widely used in semantic correspondences due to their powerful representations. Early methods formulate semantic correspondence as a geometric alignment problem, with a major focus on developing robust geometric models~\cite{hongsuck2018attentive,Rocco2017,kim2018recurrent}. Rocco~\textit{et al.}~\cite{Rocco2017,Rocco2018} propose a two-stage CNN architecture for regressing image-level transformation parameters, while other efforts regress local translation fields~\cite{kim2018recurrent,jeon2019joint,kim2019semantic}. More recent works tend to formulate semantic correspondence as a pixel-wise matching problem and cast it as a classification problem. Among these works, there are techniques focusing on developing powerful feature representations~\cite{huang2019dynamic,min2019hyperpixel,min2020learning}, correlation map filtering with 4D/6D-conv or transformers~\cite{li2020correspondence,liu2020semantic,cho2021cats,min2021convolutional,hong2022cost}, effective correspondence readout~\cite{lee2019sfnet}, and different levels of supervision~\cite{li2021probabilistic,huang2020confidence,truong2022probabilistic}. However, none of these aforementioned methods have explicitly approached the task of dense semantic correspondence from the perspective of sparse annotations.

\subsection{Teacher-Student Learning}

Teacher-student framework has been widely used in semi-supervised learning (SSL)~\cite{sohn2020fixmatch,xie2020self,tarvainen2017mean,li2022rethinking,he2022asm}, where the predictions of the teacher model on unlabeled samples serve as pseudo-labels to guide the student model.
Teacher-student framework also plays an important role in knowledge distillation~\cite{hinton2015distilling,chen2015net2net,yim2017gift,heo2019comprehensive,zhang2018deep}, where knowledge from a larger teacher model can be transferred into a smaller student model without loss of validity. Recently, Xin~\textit{et al.}~\cite{li2021probabilistic} extend teacher-student to semantic correspondence, where they distill knowledge learned from a probabilistic teacher model on synthetic data to a static student model with unlabeled real image pairs. In contrast, we directly learn from real image pairs labeled with sparse keypoints, and focus on addressing the label sparsity challenge via Siamese teacher-student network design~\cite{bromley1993signature}. Note that we tailor teacher-student learning specifically for the dense prediction task of semantic correspondence, where we conduct pixel-level semi-supervised learning within an image and generate pseudo-labels for unlabeled pixels, while most existing work focus on image-level semi-supervised learning.

\begin{figure*}[t]
	\centering
	\adjincludegraphics[width=\linewidth, trim={{0\width} {0.05\height} {0\width} {0\height}},clip]{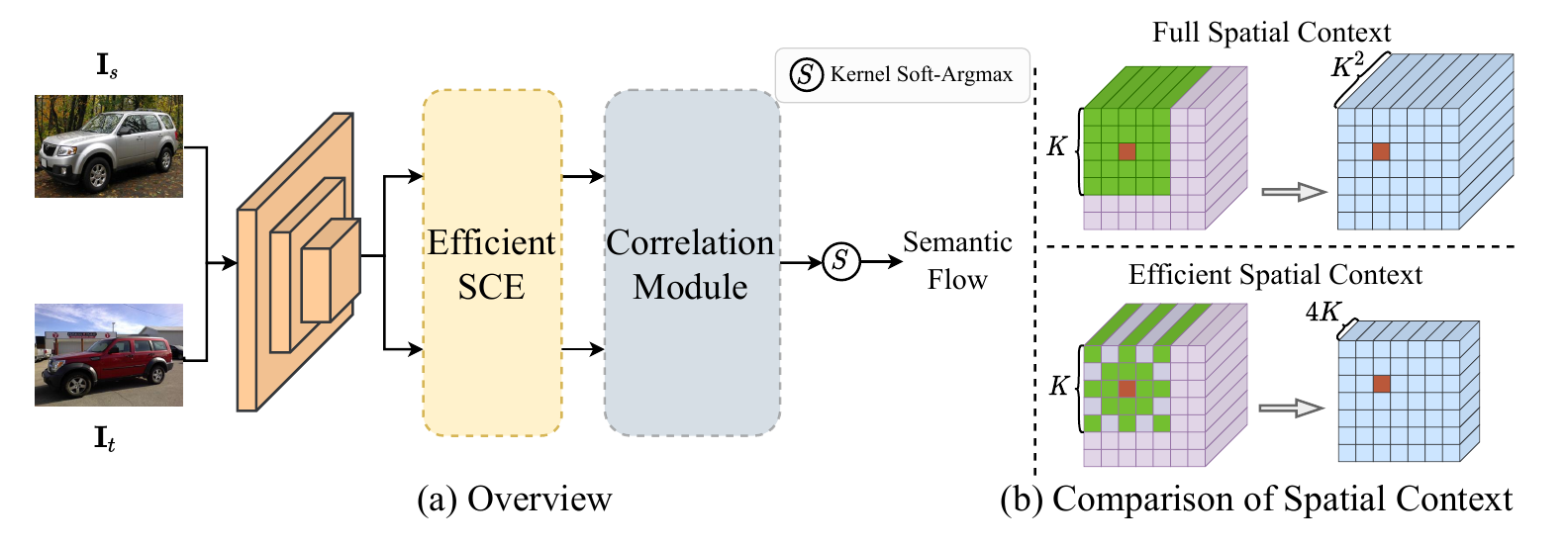}
	\caption{\small \textbf{Model overview.} (a) Illustration of our network. Our network comprises three main modules, including an efficient spatial context encoder, a correlation module, and a flow estimator. (b) Comparison between our proposed efficient spatial context and the full spatial context. Please refer Sec.~\ref{sec:model} for more details. \textit{Best viewed in color.}}
	\label{fig:model_overview}
\end{figure*}

\section{Model Architecture}
\label{sec:model}
Semantic correspondence establishes dense correspondences between a source image $\mathbf{I}_a$ and a target image $\mathbf{I}_b$. We adopt a typical CNN-based method which computes a correlation map between the convolution features of two images, based on which a dense flow field is predicted as the final output. We additionally encode spatial context efficiently to compute high-quality correlation map and develop a novel teacher-student learning strategy to cope with label sparsity.

This section introduces our simple and powerful semantic correspondence framework. As depicted in Fig.~\ref{fig:model_overview}, our framework comprises of three main modules: (1) a sparse spatial context feature extractor that encodes context information efficiently  (Sec.~\ref{subsec:model_crisscross}), (2) a correlation operator to compute the correlation map between two convolution features (Sec.~\ref{subsec:model_corr}), and (3) a flow estimation operator with high-resolution loss (Sec.~\ref{subsec:model_flow}).

\subsection{Efficient Spatial Context Encoder}
\label{subsec:model_crisscross}
Taking the conv features of the image pairs as the input, the first component of our network is an efficient spatial context encoder that incorporates spatial context into conv features. Recent methods adopt the self-similarity based descriptor to encode spatial context~\cite{huang2019dynamic,li2020correspondence}. 
However, the time complexity of the self-similarity grows quadratically with respect to the kernel size of the self-similarity descriptor due to dense sampling patterns they used~\cite{huang2019dynamic,li2020correspondence}. Inspired by the recent success of sparse attention on reducing the computational cost of non-local operation~\cite{huang2019ccnet,yue2018compact,he2020gta,zhang2019latentgnn}, we propose a spatial context encoder based on sparse sampling patterns, which efficiently encodes context information and reduces the time complexity from quadratic to linear.

As shown in Fig.~\ref{fig:model_overview}(b), at location ($i,j$), its spatial context descriptor $\mathbf{s}_{(i,j)}\in\mathbb{R}^{4K}$ is a self-similarity vector, where $K$ is the self-similarity operator kernel size. It is computed between its own feature vector $\mathbf{z}_{(i,j)}\in\mathbb{R}^{d_z}$ and its $4K$ neighboring feature vectors, where the neighbors are in its criss-cross and diagonal directions in an fixed ordered. In contrast to dense sampling patterns~\cite{huang2019dynamic,li2020correspondence}, our sparse sampling patterns reduce the time and space complexity for computing spatial descriptors from $O(K^2)$ to $O(K)$.

To combine spatial context and conv features, we employ a simple fusion step to generate the final context-aware semantic feature map $\mathbf{G}$ following~\cite{huang2019dynamic}. Concretely, we concatenate $\mathbf{z}_{(i,j)}$ and $\mathbf{s}_{(i,j)}$ and feed the result into a linear transformation with parameter $\mathbf{W}\in\mathbb{R}^{(d_z+4K)\times d_g}$ followed by a ReLU operation, resulting in a context-aware semantic feature vector $ \mathbf{g}_{(i,j)}\in\mathbb{R}^{d_g}$. 
We add subscripts to represent the context-aware semantic feature map ${\mathbf{G}}_a\in\mathbb{R}^{d_g\times h_a \times w_a}$ and ${\mathbf{G}}_b\in\mathbb{R}^{d_g\times h_b \times w_b}$ for the image $\mathbf{I}_a$ and $\mathbf{I}_b$, resp., where $h_b$, $w_b$ (resp. $h_a$, $w_a$) is the spatial size of $\mathbf{G}_{b}$ (resp. $\mathbf{G}_{a}$).

\subsection{Correlation Map Computation}
\label{subsec:model_corr}
We compute a 4D correlation map from the context-aware semantic feature maps ${\mathbf{G}}_a$, ${\mathbf{G}}_b$ and filter it with the mutual nearest neighbor module~\cite{Rocco18b}.  We denote the resulting 4D correlation map as $\mathbb{C}\in\mathcal{R}^{h_a\times w_a\times h_b\times w_b}$.

We propose to learn a high-resolution correlation map for high-quality dense matching in contrast to learning correspondence in stride16~\cite{huang2019dynamic,Rocco2018}. We upsample the correlation map $\mathbb{C}$ (4 times) instead of upsampling the feature maps for memory efficiency. We denote the resulting upsampled correlation map as
\begin{align}
    \mathbf{C} = \mathcal{U}(\mathbb{C}), \quad\mathbf{C}\in\mathbb{R}^{H_a\times W_a\times H_b \times W_b},
\end{align} where $H_a$, $W_a$, $H_b$, and $W_b$ are the upsampled spatial sizes, $\mathcal{U}$ is the upsample operation. Note that we achieve high performance with single layer feature, while DHPF~\cite{min2020learning} requires multi-layer features with higher complexity.

\subsection{Flow formation and High-resolution loss}
\label{subsec:model_flow} %
To obtain differentiable flow, we adopt the kernel soft-argmax operator~\cite{lee2019sfnet} to transform the upsampled correlation map $\mathbf{C}$ into dense semantic flow $\hat{f}$ as below:
\begin{align}
    \hat{f} = \mathcal{S}(\mathbf{C}),\quad\hat{f}\in\mathbb{R}^{2\times H_b\times W_b}
\end{align} where $\mathcal{S}$ is the kernel soft-argmax operator without any learnable parameters, $\hat{f}$ is the predicted semantic flow in the direction of the target to source.

During training, as we only have sparse keypoint labels, the ground-truth flow $f^{\text{gt}}\in\mathbb{R}^{2\times H_b\times W_b}$ have valid values only at labeled positions. We use a sparse binary label mask  $\mathbf{M}\in\mathbb{R}^{H_b\times W_b}$ to indicate valid positions with ground-truth labels as below:
\begin{equation}
\mathbf{M}(\mathbf{p}) = \left\{
\begin{array}{rcl}
1     &      & \text{if } \mathbf{p}\text{ is labeled,} \\
0       &      & \text{otherwise,}
\end{array} \right. 
\label{eq:t_i}
\end{equation} where $\mathbf{p}$ is the position index in $f^{\text{gt}}$.

Given $\mathbf{M}$, the objective is then defined as the L2 norm between the predicted flow and the ground-truth flow at labeled subpixel positions:
\begin{align}
	&L^{\text{gt}}(\mathbf{p}) = \|\hat{f}(\mathbf{p})-f^{\text{gt}}(\mathbf{p})\|_2\cdot \mathbf{M}(\mathbf{p})
\end{align} where $L^{\text{gt}}(\mathbf{p})$ is the ground-truth loss at position $\mathbf{p}$. It is worth noting that our network does not involve any 4D-conv or transformer for correlation refinement~\cite{cho2021cats,huang2019dynamic}, but as shown later it achieves high performance thanks to our efficient spatial context encoder and high-resolution design.

\begin{figure}[t]
	\centering
	\adjincludegraphics[width=\linewidth, trim={{0.08\width} {0} {0.02\width} {0\height}},clip]{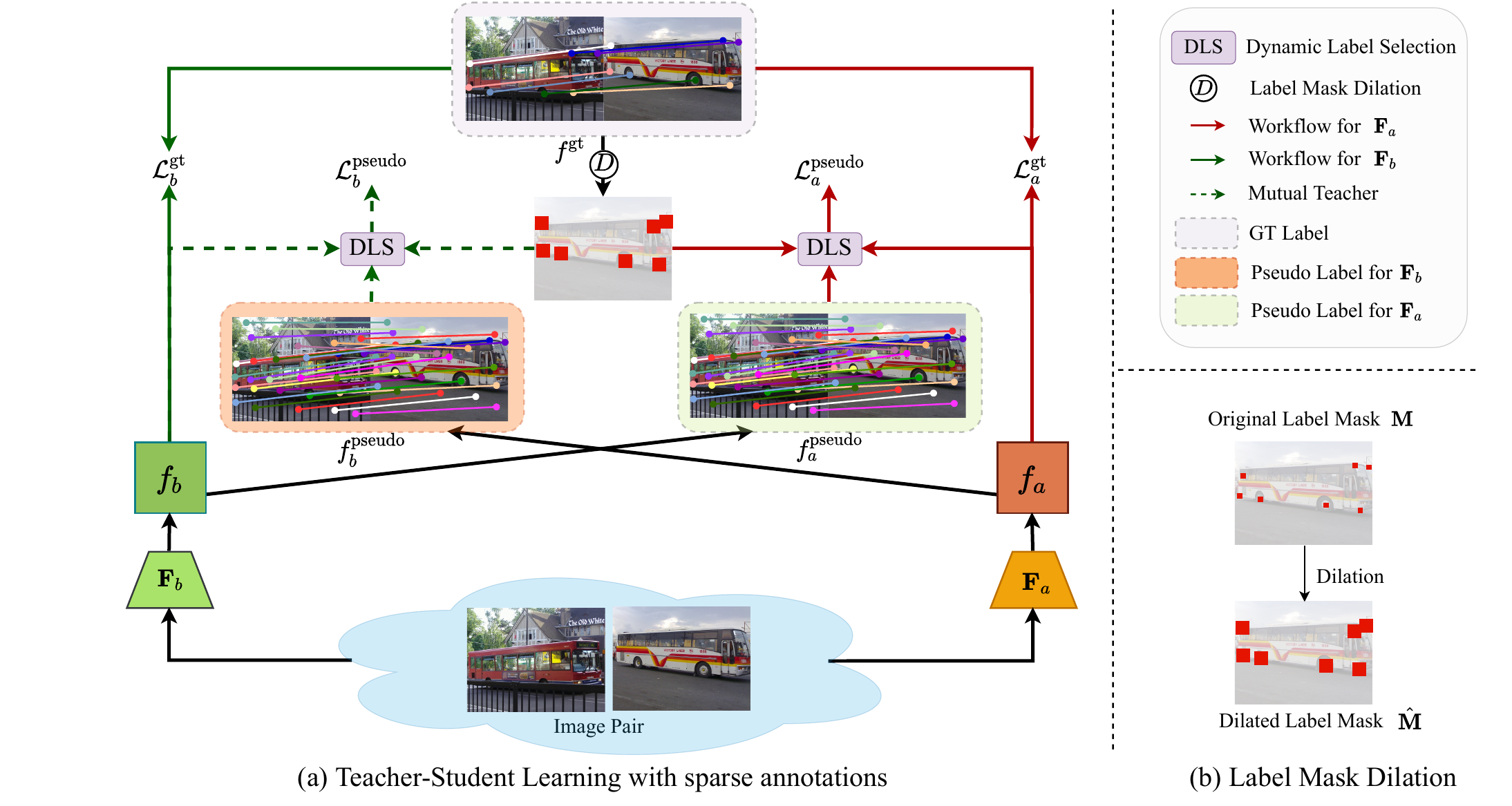}
    \caption{\textbf{Sparse Label Densification with Teacher-Student Learning.} (a) Our Teacher-Student Learning Pipeline. Solid lines stand for Single Offline Teacher, with additional dashed lines standing for Mutual Online Teacher. (b) Illustration of Label Mask Dilation. Please refer to Sec.~\ref{sec:method} for more details. \textit{Best viewed in color.}}
	\label{fig:pipeline}
\end{figure}

\section{Learning with Sparse Annotations}

\label{sec:method}

While our network design enables us to encode spatial context efficiently and utilize high-resolution correlation maps, the sparsely-annotated keypoint pairs ($8$ on average on PF-PASCAL~\cite{ham2018proposal}) greatly hinder the learning of the dense matching model. We address this with a novel teacher-student learning framework which we will elaborate below.

Our goal is to enrich the supervision when only sparse annotations are provided, as shown in Fig.~\ref{fig:pipeline}. We first densify the sparse labels with a teacher-student paradigm (Sec.~\ref{sec:sub_tea_stu}). Then we introduce two novel techniques to denoise the generated pseudo-labels (Sec.~\ref{sec:label_generation}): (a) leveraging spatial-priors and (b) loss-driven dynamic label selection. Finally, we investigate two variants of the proposed learning paradigm (Sec.~\ref{sec:var}).

\subsection{Sparse Label Densification via Teacher-Student Learning}
\label{sec:sub_tea_stu}
To enrich the sparse supervision signals, we generate dense pseudo-labels for unlabeled region via a teacher-student paradigm, which consists of a student model $\mathbf{F}_s$ and a teacher model $\mathbf{F}_t$. The teacher model $\mathbf{F}_t$ trained with sparse annotations generates dense flows $\hat{f}_t$, providing pseudo-labels $f_s^{\text{pseudo}}$ for the student model $\mathbf{F}_s$. Formally,
\begin{align}
    f_s^{\text{pseudo}}=\mathbf{F}_t(\mathbf{I}_a,\mathbf{I}_b), \quad f_s^{\text{pseudo}}\in\mathbb{R}^{2\times H_b\times W_b}.
\end{align} Then, the optimization objective $L_s(\mathbf{p})$ for the student $\mathbf{F}_s$ is a combination of the ground-truth loss $L_s^{\text{gt}}(\mathbf{p})$ and a dense pseudo-label loss $L_s^{\text{pseudo}}(\mathbf{p})$ calculated as follows:
\begin{align}
    &L_s(\mathbf{p}) = L_s^{\text{gt}}(\mathbf{p}) + \lambda L_s^{\text{pseudo}}(\mathbf{p}) \\
           & L_s^{\text{gt}}(\mathbf{p})=\|\hat{f}_s(\mathbf{p})-f^{\text{gt}}(\mathbf{p})\|_2\cdot\mathbf{M}(\mathbf{p}) \\
       & L_s^{\text{pseudo}}(\mathbf{p})=\|\hat{f}_s(\mathbf{p})-f_s^{\text{pseudo}}(\mathbf{p})\|_2
\end{align} where $\hat{f}_s\in\mathbb{R}^{2\times H_t\times W_t}$ is the predicted flow of the student model $\mathbf{F}_s$ given $\mathbf{I}_a$ and $\mathbf{I}_b$, $\lambda$ is the scale hyper-parameter, $\mathbf{p}$ indexes the positions in $\hat{f}_s$.

\subsection{High Quality Pseudo-label Generation}
\label{sec:label_generation}
The dense pseudo-labels generated by the teacher model are inevitably unreliable and inaccurate for supervision. To filter out erroneous pseudo-labels, we use: (a) label filtering based on spatial priors, and (b) loss-driven dynamic label selection.

\subsubsection{Spatial-prior Based Label Filtering.}
Our key insight is that, as the annotated keypoints are in the object foreground region, we are able to suppress noisy background pseudo-labels by exploiting the spatial-smoothness prior of the semantic correspondence in the neighborhood of the sparse keypoints. Motivated by this, we generate a densified binary label mask $\hat{\mathbf{M}}$ via dilating the sparse label mask $\mathbf{M}$ as follows, which will be used for label filtering:
\begin{align}
    &\bar{\mathbf{M}} = \mathbf{M} * \mathcal{K}\\
    \label{eq:dilate}
    &\hat{\mathbf{M}}(\mathbf{p}) = \left\{
\begin{array}{rcl}
1     &      & \text{if } \bar{\mathbf{M}}(\mathbf{p})>0 \\
0       &      & \text{otherwise}
\end{array} \right. 
\end{align} where $*$ is a convolution operator with zero padding, $\mathcal{K}\in\mathcal{R}^{k\times k}$ is a kernel filled with one with $k$ as the kernel size. Note that dilation here refers to expanding the existing foreground region in $\mathbf{M}$. Compared with using CAM~\cite{zhou2016learning} or uncertainty estimation~\cite{li2021probabilistic}, our proposed label filtering technique is easy to implement and utilizes the spatial prior around the sparse annotations.

Given the dilated label mask $\hat{\mathbf{M}}$, the pseudo-loss $\hat{L}_s^{\text{pseudo}}(\mathbf{p})$ for the student model is calculated as below:
\begin{align}
    &\hat{L}_s^{\text{pseudo}}(\mathbf{p})=\|\hat{f}_s(\mathbf{p})-f_s^{\text{pseudo}}(\mathbf{p})\|_2\cdot\hat{\mathbf{M}}(\mathbf{p})\label{eq:la_final}.
\end{align} where $\mathbf{p}$ indexes the positions. In this way, we are able to significantly suppress noisy background pseudo-labels as shown in Sec~\ref{sec_exp:ablation}.

\subsubsection{Loss-driven Dynamic Label Selection.}
While many background pseudo-labels can be filtered out by our dilated label mask, some noisy labels still exist due to inaccurate predictions from the teacher model. To further filter out inaccurate labels, we introduce a loss-driven label selection strategy following the small-loss principle~\cite{han2018co}.
Denoting $R$ as the ratio of pixels being selected, we choose the pixel set $\mathcal{P}$ on the foreground region in $\hat{\mathbf{M}}$ with the smallest loss as below:
\begin{align}
&\mathcal{P} = \argmin_{\bar{\mathcal{D}}: |\bar{\mathcal{D}}|\geq R(T)N_{\hat{\mathbf{M}}} \wedge \bar{\mathcal{D}}\subseteq \hat{\mathcal{D}}}\sum_{\mathbf{p}\in\bar{\mathcal{D}}} \hat{L}_s^{\text{pseudo}}(\mathbf{p})\\
&\hat{\mathcal{D}} = \{\mathbf{p}\mid\hat{\mathbf{M}}(\mathbf{p})=1\}
\label{eq:D_a}
\end{align} where $R(T)$ controls the selection percentage in training epoch $T$, $\mathbf{p}$ indexes the positions, $\hat{\mathcal{D}}$ is a collection of foreground positions in the dilated label mask $\hat{\mathbf{M}}$, $N_{\hat{\mathbf{M}}}$ is the total number of non-zero positions in $\hat{\mathbf{M}}$.

Hence the final optimization objective $\mathcal{L}_s$ for the student model $\mathbf{F}_s$ over an image pair is a combination of the sparse ground-truth loss $\mathcal{L}_s^{\text{gt}}$ at labeled positions $\mathcal{G}=\{\mathbf{p}\mid\mathbf{M}(\mathbf{p})=1\}$ and the dense pseudo loss $\mathcal{L}_s^{\text{pseudo}}$ at selected positions $\mathcal{P}$ as below:
\begin{align}
  &\mathcal{L}_s = \mathcal{L}_s^{\text{gt}} + \lambda\mathcal{L}_s^{\text{pseudo}}\\
&\mathcal{L}_s^{\text{pseudo}}=\frac{1}{|\mathcal{P}|}\sum_{\mathbf{p}\in \mathcal{P}}^{}\hat{L}_s^{\text{pseudo}}(\mathbf{p})\\
&\mathcal{L}_s^{\text{gt}}=\frac{1}{|\mathcal{G}|}\sum_{\mathbf{p}\in \mathcal{G}}^{}L_s^{\text{gt}}(\mathbf{p})
\label{eq:loss_model_a}
\end{align}

\subsection{Variants of Teacher-Student Learning}
\label{sec:var}
To investigate the optimization strategy of our proposed teacher-student learning, we propose two variants of our learning strategy, including (a) single offline teacher, and (b) mutual online teachers, which are detailed below.

\subsubsection{Single offline Teacher (ST).} This variant consists of two learning stages. Specifically, we first learn a baseline network, which acts as the teacher model, given spare ground-truth annotation only. In the second stage, the pseudo-labels are generated by the fixed teacher network as described in Sec.~\ref{sec:sub_tea_stu} and Sec.~\ref{sec:label_generation}. Given the enriched supervision, we then train the student model from scratch, which is used for the inference stage finally. 

\subsubsection{Mutual online Teacher (MT).} Inspired by the recent advances in multi-view learning~\cite{blum1998combining,yang2021deep}, we additionally explore a one-stage variant with two mutual online teachers which learn from scratch. We simultaneously train two networks of the same architecture, each of which takes predictions from the other network as the pseudo-labels for optimization. These two networks can learn knowledge of correspondence with enriched pseudo-labels from each other. The one with a higher validation performance is selected for the inference stage.

Specifically, we maintain two networks $\mathbf{F}_s$ and $\mathbf{F}_t$ of the same architecture. The network $\mathbf{F}_t$ (resp. $\mathbf{F}_s$) provides its predicted flow $\hat{f}_t$ (resp. $\hat{f}_s$) as the pseudo-label $f_s^{\text{pseudo}}$ (resp. $f_t^{\text{pseudo}}$) for the peer network $\mathbf{F}_s$ (resp. $\mathbf{F}_t$). Both networks use the shared dilated label mask $\hat{\mathbf{M}}$ for label filtering. For each model, the pseudo-loss filtered by dilated label masks is described as below:
\begin{align}
    \hat{L}_s^{\text{pseudo}}(\mathbf{p})
    &=  \|\hat{f}_s(\mathbf{p})-f_s^{\text{pseudo}}(\mathbf{p})\|_2\cdot\hat{\mathbf{M}}(\mathbf{p})\\
    \hat{L}_t^{\text{pseudo}}(\mathbf{p})
    &=  \|\hat{f}_t(\mathbf{p})-f_t^{\text{pseudo}}(\mathbf{p})\|_2\cdot\hat{\mathbf{M}}(\mathbf{p}),
\end{align} where $\mathbf{p}$ indexes the position, $f_s^{\text{pseudo}}$ (resp. $f_t^{\text{pseudo}}$) equals to $\hat{f}_t$ (resp. $\hat{f}_s$). $\hat{L}_s^{\text{pseudo}}(\mathbf{p})$ and $\hat{L}_t^{\text{pseudo}}(\mathbf{p})$ will then go through the dynamic label selection procedure as described in Sec.~\ref{sec:label_generation} to compute pseudo-label loss $\mathcal{L}_s^{\text{pseudo}}$ and $\mathcal{L}_t^{\text{pseudo}}$, respectively. The final optimization objective for each model is a combination of sparse ground-truth loss and pseudo loss as below:
\begin{align}
  &\mathcal{L}_s = \mathcal{L}_s^{\text{gt}} + \lambda\mathcal{L}_s^{\text{pseudo}}\\
  &\mathcal{L}_t = \mathcal{L}_t^{\text{gt}} + \lambda\mathcal{L}_t^{\text{pseudo}}
\end{align}

\section{Experiments}

We evaluate our method on the supervised semantic correspondence task by conducting comprehensive experiments on three public benchmarks: PF-PASCAL~\cite{ham2018proposal}, PF-WILLOW~\cite{ham2016proposal}, and SPair-71k~\cite{min2019spair}.
In the following sections, we first elaborate on the implementation details of our proposed method in Sec.~\ref{sec_exp:detail}, and follow that with the quantitative and qualitative comparison with prior state-of-the-art (SOTA) competitors in Sec.~\ref{sec_exp:quantitative}. Then, we provide ablation studies and comprehensive analysis in Sec.~\ref{sec_exp:ablation}. For more detailed results and analysis, we refer readers to the supplementary material.

\begin{table*}[t]
	\centering
	\footnotesize
	\renewcommand{\arraystretch}{1.1}
    \renewcommand{\tabcolsep}{2pt}
	\caption{\textbf{Comparison with SOTA methods on SPair-71k~\cite{min2019spair}}. Per-class and overall PCK ($\alpha_\text{bbox}=0.1$) results are shown in the table. Numbers in bold indicate the best performance and underlined ones are the second best. All models in this table use ResNet101 as the backbone. \textit{Sup.} denotes the type of supervision. * means the backbone is finetuned. $\dagger$ means ground truth bbox used.}
	
	\resizebox{1.0\textwidth}{!}{
		\begin{tabular}{@{}llccccccccccccccccccc@{}}
			\toprule
			Sup.  & Methods  & aero & bike &bird & boat &bottle& bus & car &cat& chair&cow&dog&horse&mbike&person&plant&sheep&train&tv&\textbf{all}  \\  

			\midrule
			\multirow{2}{*}{self} & CNNGeo~\cite{Rocco2017}             & 23.4 &16.7 &40.2& 14.3& 36.4 &27.7 &26.0& 32.7 &12.7 &27.4& 22.8& 13.7 &20.9 &21.0 &17.5& 10.2 &30.8& 34.1 &20.6\\
			& A2Net~\cite{hongsuck2018attentive} & 22.6  &18.5  &42.0 & 16.4 & 37.9  &30.8  &26.5  &35.6 & 13.3 & 29.6  &24.3  &16.0  &21.6  &22.8  &20.5  &13.5  &31.4 & 36.5 & 22.3\\
			\midrule
			\multirow{2}{*}{weak} &WeakAlign~\cite{Rocco2018}         & 22.2 &17.6& 41.9 &15.1 &38.1 &27.4 &27.2& 31.8& 12.8& 26.8& 22.6 &14.2 &20.0 &22.2 &17.9 &10.4& 32.2 &35.1 &20.9 \\
			&NCNet~\cite{Rocco18b}              & 17.9 & 12.2 & 32.1 & 11.7&  29.0 & 19.9 & 16.1 & 39.2&  9.9&  23.9&  18.8&  15.7&  17.4&  15.9&  14.8 & 9.6&  24.2&  31.1&  20.1\\
			\midrule
			trn-none / &HPF~\cite{min2019hyperpixel}   &25.2 &18.9 &52.1 &15.7& 38.0& 22.8 &19.1 &52.9 &17.9& 33.0 &32.8 &20.6& 24.4 &27.9 &21.1 &15.9& 31.5& 35.6 &28.2\\
			val-strong &SCOT~\cite{liu2020semantic}   &34.9 &20.7 &63.8 &21.1& 43.5& 27.3 &21.3 &63.1 &20.0& 42.9 &42.5 &31.1& 29.8 &35.0 &27.7 &24.4& 48.4& 40.8 &35.6\\
			\cmidrule{1-21}
				\multirow{8}{*}{strong}&DHPF~\cite{min2020learning}  &38.4  &23.8 & 68.3  &18.9 & 42.6 & 27.9 & 20.1  &61.6  &22.0 & 46.9 & 46.1 & 33.5 & 27.6 & 40.1  &27.6 &28.1 &49.5  &46.5 & 37.3\\
		&PMD~\cite{li2021probabilistic}&38.5& 23.7& 60.3& 18.1& 42.7& 39.3& 27.6& 60.6& 14.0& 54.0& 41.8& 34.6& 27.0& 25.2& 22.1& 29.9& 70.1& 42.8& 37.4\\
		
		&MMNet*~\cite{zhao2021multi}   &43.5& 27.0& 62.4& 27.3& 40.1& 50.1& 37.5& 60.0& 21.0& 56.3& 50.3& 41.3& 30.9& 19.2& 30.1& 33.2& 64.2& 43.6& 40.9\\
	
		&CHM~\cite{min2021convolutional}&49.6 &29.3& 68.7& 29.7& 45.3& 48.4& 39.5& 64.9& 20.3& 60.5& 56.1& 46.0& 33.8& 44.3& 38.9& 31.4& 72.2& 55.5& 46.3\\
		
		&$\text{CATs}^{\dagger}$*~\cite{cho2021cats}   &52.0 &34.7& 72.2& 34.3& 49.9& \underline{57.5}& \underline{43.6}& 66.5& \underline{24.4}& 63.2& 56.5& \textbf{52.0}& 42.6& 41.7& 43.0& 33.6& 72.6& 58.0& 49.9\\
		
		&PMNC*~\cite{lee2021patchmatch}   &54.1& 35.9& 74.9& \underline{36.5}& 42.1& 48.8& 40.0& 72.6& 21.1& 67.6& 58.1& \underline{50.5}& 40.1& \textbf{54.1}& 43.3& 35.7& 74.5& 59.9& 50.4\\
		
		\cmidrule{2-21}

		& \textbf{Ours (ST)*} &\underline{56.9} &\underline{37.0} &\underline{76.2} &33.9 &\underline{50.1} &51.7 &42.4 &\underline{68.2}&22.4 &\underline{70.7}&\underline{61.0} &47.7&\underline{43.6} &47.8&\underline{47.8} &\underline{38.6}&\underline{77.0} &\underline{67.1} & \underline{52.4}\\
		& \textbf{Ours (MT)*} &\textbf{57.1} &\textbf{40.3} &\textbf{78.3} &\textbf{38.1} &\textbf{51.8} &\textbf{57.8} &\textbf{47.1} &\textbf{67.9}&\textbf{25.2} &\textbf{71.3}&\textbf{63.9} &49.3&\textbf{45.3} &\underline{49.8}&\textbf{48.8} &\textbf{40.3}&\textbf{77.7} &\textbf{69.7} & \textbf{55.3}\\
			\bottomrule
	\end{tabular}}

	\label{tab:Spair}
\end{table*}

\subsection{Implementation Details}\label{sec_exp:detail}
\subsubsection{Datasets.}

\textit{SPair-71k} is a newly-released challenging and largest-scale benchmark~\cite{min2019spair}. There are keypoint-annotated 70,958 image pairs with large viewpoint and scale variation in diverse scenes. SPair-71k~\cite{min2019spair} is a reliable test bed for studying real problems of semantic matching. 
\textit{PF-PASCAL} dataset~\cite{ham2018proposal} contains 1351 image pairs with limited variability and scale, which is approximately split into 700, 300, and 300 pairs for train, val, and test set, resp.
\textit{PF-WILLOW}~\cite{ham2016proposal} dataset consists of 900 image pairs of 4 categories, which is a widely-used benchmark for the verification of generalization ability.

\subsubsection{Evaluation Metric.}
In line with prior work, we report the percentage of correct keypoints (PCK)~\cite{yang2013articulated}. The predicted keypoints are considered to be correct if they lie within $\alpha\cdot\max(h,w)$ pixels from the ground-truth keypoints for $\alpha\in[0,1]$, where $h$ and $w$ are the height
and width of either an image ($\alpha_\text{img}$) or an object bounding box ($\alpha_\text{bbox}$). 

\subsubsection{Experimental Configuration.}
For the feature extractor, we use ResNet-101~\cite{he2016deep} pre-trained on ImageNet with a single feature at stride 16. Learnable parameters are randomly initialized. For our base model, we set Efficient SCE kernel size $K=7$ and $d_g=2048$ for SPair-71k; $K=13$ and $d_g=1024$ for PF-PASCAL, resp. We upsample the correlation map to stride 4 for high-resolution loss. For label mask dilation, dilation kernel size $k=7$ is set for both SPair-71k and PF-PASCAL by validation search. For dynamic label selection, we set $R(T)$ linearly increases from the ratio of $20\%$ to $90\%$ in a duration of 10 epochs for both SPair-71k and PF-PASCAL. $\lambda$ is 10.0 for weighting pseudo-loss. We strictly follow previous work for data augmentation~\cite{cho2021cats} (e.g., color jittering) except that~\cite{cho2021cats} uses ground truth box for random crop while we do not. An AdamW optimizer with a learning rate of 3e-6 for the backbone and 3e-5 for the remaining parameters are used. All the implementations are in PyTorch~\cite{paszke2017automatic}.

Images of all three datasets are resized to 256$\times$256. Our model is trained on PF-PASCAL and SPair-71k, resp. Following the previous work~\cite{min2020learning,cho2021cats,huang2019dynamic}, we validate the generalization ability of our method by testing on PF-WILLOW with our model trained on PF-PASCAL without any finetuning.

\begin{table}[t]
	\centering
	\footnotesize
	\caption{\textbf{Comparison with SOTA methods on PF-PASCAL}~\cite{ham2018proposal}. Numbers in bold indicate the best performance and underlined ones are the second best. $\dagger$ means ground truth bbox used.}
	\renewcommand{\arraystretch}{1.1}
    \renewcommand{\tabcolsep}{3pt}
	\resizebox{0.75\linewidth}{!}{
		\begin{tabular}{@{}llc c c c@{}}
		\toprule
		\multirow{2}{*}{Sup.}  &	\multirow{2}{*}{Methods}  & \multicolumn{3}{c}{\textbf{PCK}@$\alpha_\text{img}$} &$\alpha_\text{bbox}$\\  
			\cmidrule(lr){3-5}
			\cmidrule{6-6}
			&& $\alpha=0.05$&$\alpha=0.10$&$\alpha=0.15$ & $\alpha=0.1$\\
			\midrule
	
			none&PF-LOM{\scriptsize{HOG}}~\cite{ham2018proposal} & 31.4&62.5&79.5  & 45.0 \\
			\midrule
			self&CNNGeo{\scriptsize{ResNet-101}}~\cite{Rocco2017} &  41.0 & 69.5 & 89.4 & 68.0 \\
\midrule
			\multirow{4}{*}{weak}&WeakAlign{\scriptsize{ResNet-101}}~\cite{Rocco2018}   &49.0&74.8&84.0 & 72.0\\
			&NC-Net{\scriptsize{ResNet-101}}~\cite{Rocco18b}   &54.3&78.9&86.0 & 70.0\\
			&DCCNet{\scriptsize{ResNet-101}}~\cite{huang2019dynamic}          &55.6&82.3&90.5 & -\\
			&GSF{\scriptsize{ResNet-101}}~\cite{jeon2020guided}         &62.8&84.5&93.7 & -\\
		
		\midrule	
			\multirow{1}{*}{trn-none}&HPF{\scriptsize{ResNet-101}}~\cite{min2019hyperpixel}      &60.1 & 84.8 & 92.7 & 78.5 \\
			val-strong&SCOT{\scriptsize{ResNet-101}}~\cite{liu2020semantic}           &63.1 & 85.4 & 92.7&-\\
		\midrule
			\multirow{10}{*}{strong}&SCNet{\scriptsize{VGG-16}}~\cite{khan2017} & 36.2&72.2&82.0 & 48.2\\
			&ANCNet{\scriptsize{ResNet-101}}~\cite{li2020correspondence}           & - & 86.1 & - & -\\
			&DHPF{\scriptsize{ResNet-101}}~\cite{min2020learning}  & 75.7 & 90.7 & 95.0 & 87.8\\
			&PMD{\scriptsize{ResNet-101}}~\cite{li2021probabilistic}  & - & 90.7 & - & -\\
			&MMNet{\scriptsize{ResNet-101}}~\cite{zhao2021multi}  & 77.6 & 89.1 & 94.3 & -\\
			&CHM{\scriptsize{ResNet-101}}~\cite{min2021convolutional}  & 80.1 & 91.6 & - & -\\
			&$\text{CATs}^{\dagger}${\scriptsize{ResNet-101}}~\cite{cho2021cats}  & 75.4 & 92.6 & \underline{96.4} & 89.2\\
			&PMNC{\scriptsize{ResNet-101}}~\cite{lee2021patchmatch}  & \textbf{82.4} & 90.6 & - & -\\
			\cmidrule{2-6}
			&\textbf{Ours (ST)}{\scriptsize{ResNet-101}}           &81.4 &\underline{92.9}&96.1 & \underline{90.5}\\
			&\textbf{Ours (MT)}{\scriptsize{ResNet-101}}           &\underline{81.5} &\textbf{93.3}&\textbf{96.6} & \textbf{91.2}\\
			\bottomrule
	\end{tabular}}
	\label{tab:evalpfpascal_three}
\end{table}

\subsection{Comparison with State-of-the-art Methods}\label{sec_exp:quantitative}
\begin{figure}[t]
	\centering
	\adjincludegraphics[width=\linewidth, trim={{0\width} {0.0\height} {0\width} {0.0\height}},clip]{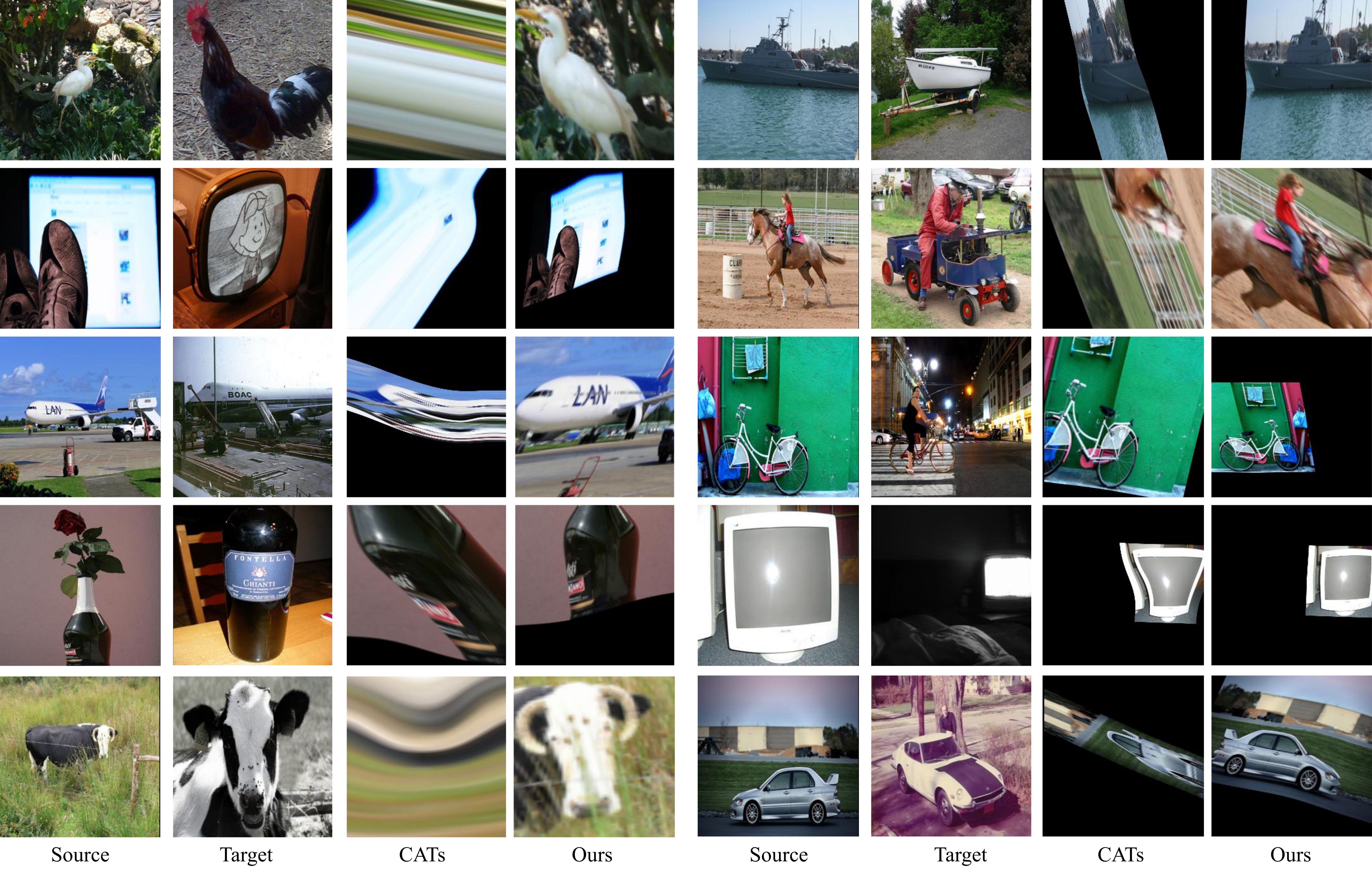}

     \caption{\textbf{Qualitative results of our method on SPair-71k~\cite{min2019spair}.} From left to right are source image, target image, result from CATs~\cite{cho2021cats}, and result from ours (MT), resp.}
	\label{fig:vis_spair}
\end{figure}

\subsubsection{SPair-71k.}
We compare our method with the most recent  work~\cite{min2019hyperpixel,liu2020semantic,min2020learning,zhao2021multi,lee2021patchmatch,cho2021cats} on SPair-71k in Table~\ref{tab:Spair}. Our two variant settings (ST and MT) both achieve an overall SOTA results, with our method (MT) achieving an overall PCK ($\alpha_{\text{img}}=0.1$) of $55.3\%$, outperforming the previous SOTA~\cite{lee2021patchmatch} by a large margin ($4.9\%$). Note that our method does not involve any parameterized correlation refinement compared with~\cite{lee2021patchmatch}, clearly illustrating the power of the proposed pipeline. Fig.~\ref{fig:vis_spair} shows qualitative results on SPair-71k. We observe that our method is robust to diverse variations in scale and viewpoint thanks to our enriched training signals.
\subsubsection{PF-PASCAL.}
\label{sec:pf-pascal-benchmark}
Our results on PF-PASCAL are summarized in Table~\ref{tab:evalpfpascal_three}. Our method outperforms the previous state-of-the-art~\cite{lee2021patchmatch,cho2021cats} on 
almost all thresholds even if the performance on PF-PASCAL is near saturated, reaching a new SOTA of $93.3\%$ PCK ($\alpha=0.1$). Note that even if we did not use sophisticated parameterized correlation map refinement as in PMNC~\cite{lee2021patchmatch}, we can still achieve comparable PCK at $\alpha=0.05$.

\subsubsection{PF-WILLOW.} We test on PF-WILLOW~\cite{ham2016proposal} using our model trained on PF-PASCAL~\cite{ham2018proposal} to verify dataset generalization ability of our method. As shown in Table~\ref{tab:evalwillow}, our method (MT) outperforms the prior SOTA~\cite{min2021convolutional,cho2021cats} in $\alpha=0.05$, $0.1$ by $1.4\%$ and $0.6\%$, resp, indicating superior dataset generalization ability of our learning method. Note that our method (MT) are $0.5\%$ behind at $\alpha=0.15$ compared with~\cite{cho2021cats}, we argue that CATs~\cite{cho2021cats} used ground truth bounding box during training while we did not.

\begin{table}[t]
	\centering
	\footnotesize
	\renewcommand{\arraystretch}{1.1}
    \renewcommand{\tabcolsep}{3pt}
	\caption{\textbf{Comparison with SOTA methods on PF-WILLOW~\cite{ham2016proposal}.} Numbers in bold indicate the best performance and underlined ones are the second best. $\dagger$ means ground truth bbox used.}
	\resizebox{0.75\linewidth}{!}{
		\begin{tabular}{@{}llccc@{}}
			\toprule
			\multirow{2}{*}{Sup.} & \multirow{2}{*}{Methods} & \multicolumn{3}{c}{\textbf{PCK}@$\alpha_\text{bbox}$}  \\  
			\cmidrule{3-5}
			&&$\alpha=0.05$&$\alpha=0.10$&$\alpha=0.15$ \\
			\midrule
			none&PF-LOM{\scriptsize{HOG}}~\cite{ham2018proposal} &28.4&56.8&68.2  \\
			\midrule
	self&CNNGeo{\scriptsize{ResNet-101}}~\cite{Rocco2017}&36.9&69.2&77.8\\
	\midrule
			\multirow{4}{*}{weak}&WeakAlign{\scriptsize{ResNet-101}}~\cite{Rocco2018} &37.0&70.2&79.9\\
			&NC-Net{\scriptsize{ResNet-101}}~\cite{Rocco18b}  &44.0&72.7&85.4\\
			&DCCNet{\scriptsize{ResNet-101}}~\cite{huang2019dynamic}     &43.6&73.8&86.5\\
			&GSF{\scriptsize{ResNet-101}}~\cite{jeon2020guided}  &47.0&75.8&88.9\\
			\midrule
			\multirow{1}{*}{trn-none}&HPF{\scriptsize{ResNet-101}}~\cite{min2019hyperpixel} & 45.9 &74.4 &85.6\\

			val-strong&SCOT{\scriptsize{ResNet-101}}~\cite{liu2020semantic}          &47.8 &76.0& 87.1\\
			\midrule
			\multirow{7}{*}{strong}&SCNet{\scriptsize{VGG-16}}~\cite{khan2017}&38.6&70.4&85.3\\
			&DHPF{\scriptsize{ResNet-101}}~\cite{min2020learning}& 49.5 & 77.6 & 89.1\\
		
			&PMD{\scriptsize{ResNet-101}}~\cite{li2021probabilistic}& - & 75.6 & -\\
		
			&CHM{\scriptsize{ResNet-101}}~\cite{min2021convolutional}  & 52.7 & \underline{79.4} & -\\
			&$\text{CATs}^{\dagger}${\scriptsize{ResNet-101}}~\cite{cho2021cats}&50.3& 79.2& \textbf{90.3}\\

			\cmidrule{2-5}
			&\textbf{Ours (ST)}\scriptsize{ResNet-101}      &\underline{53.5}&\underline{79.4}&89.5\\
			&\textbf{Ours (MT)}\scriptsize{ResNet-101}      &\textbf{54.1}&\textbf{80.0}&\underline{89.8}\\

			\bottomrule
	\end{tabular}}

	\label{tab:evalwillow}
\end{table}

\begin{table}[t]
\footnotesize
\renewcommand{\arraystretch}{1.2}
\renewcommand{\tabcolsep}{6pt}
\caption{\textbf{Effects of each component on SPair-71K~\cite{min2019spair} test split.} HRLoss refers to high-resolution loss, FT refers to finetuning the backbone, Teacher-Student here refers to the variant with mutual online teacher.}
	\centering
	\resizebox{1.0\linewidth}{!}{
		\begin{tabular}{@{}cccccccc@{}}
			\toprule                
			
			\multirow{2}{*}{Model ID} & \multirow{2}{*}{Efficient-SCE} & \multirow{2}{*}{HRLoss}& \multirow{2}{*}{Finetune}& \multirow{2}{*}{Teacher-Student} & \multicolumn{2}{c}{Label Denoise}& \multirow{2}{*}{~PCK~}  \\  
			\cmidrule{6-7}
			&&&&& Dynamic Selection & Mask Dilation \\
			\midrule
			A0&- & -  & -& -& -  & - &   14.7      \\
			\midrule
			A1&\cmark\quad &  - & -  & -&- & - &  33.4          \\
			A2&\cmark\quad &  \cmark\quad & -  &-& - & - &  40.6  \\
			A3&\cmark\quad &  \cmark\quad& \cmark\quad  & -&- & - &  49.8  \\
			A4&\cmark\quad &  \cmark\quad& \cmark\quad  & \cmark&- & - &  49.7 \\
			A5&\cmark\quad & \cmark\quad& \cmark\quad  &\cmark&\cmark\quad& - &  51.5  \\
			A6&\cmark\quad & \cmark\quad& \cmark\quad  &\cmark&\cmark\quad& \cmark\quad &  \textbf{55.3}  \\
			\bottomrule
		\end{tabular}
	}
	\label{tab:ablation_ind}
\end{table}

\subsection{Ablation Study}
\label{sec_exp:ablation}
In this section, we conduct ablation studies to verify the effectiveness of each individual module of the proposed model.
We train all the variants on the training split of SPair-71k~\cite{min2019spair} and report PCK ($\alpha_{\text{bbox}}=0.1$) on the test split. Each ablation experiment is conducted under the same experimental setting for a fair comparison.

\subsubsection{Effect of individual modules.}
Table~\ref{tab:ablation_ind} summarizes the ablation results of each individual module. First, we note that applying the proposed Efficient-SCE (ID A1) yields significant gain over the baseline (ID A0), showing the effectiveness of the proposed feature enhancement module. Second, enforcing high-resolution loss improves to a remarkable $49.8\%$ after finetuning (ID A3). Our proposed network achieves competitive results without any Conv4D or transformer modules for correlation map refinement. Third, densifying labels combined with our two denoising techniques achieves $5.5\%$ boost further and promotes the performance to $55.3\%$, showing the effectiveness of our proposed learning strategy. In contrast, teacher-student learning alone without any denoising provides little boost as the dense pseudo-labels might be too noisy (ID A4).

\begin{table}[t]
\begin{minipage}{.5\textwidth}
\renewcommand{\tabcolsep}{6pt}
\renewcommand{\arraystretch}{1.2}
\caption{\textbf{Comparing single offline teacher and mutual online teacher} setting on SPair-71K~\cite{min2019spair} test split.}
	\centering
	\resizebox{0.65\linewidth}{!}{
		\begin{tabular}{@{}ll@{}}
			\toprule                
			Variant Setting &  PCK                    \\
			\midrule
			none & 49.8      \\
			single offline teacher & 52.4      \\
			mutual online teacher & \textbf{55.3}      \\

			\bottomrule
		\end{tabular}
	}
	\label{tab:ablation_setting}
\end{minipage}
\hfill
\begin{minipage}{.45\textwidth}
\renewcommand{\tabcolsep}{6pt}
\caption{\textbf{Effects of kernel size} for label mask dilation on SPair-71K~\cite{min2019spair} test split.}
	\centering
	\resizebox{0.65\linewidth}{!}{
		\begin{tabular}{@{}cc@{}}
			\toprule                
			Dilation Kernel Size &  PCK \\
			\midrule
			none & 49.8      \\
			3& 54.0      \\
			7& \textbf{55.3}      \\
			15& 52.0      \\
			\bottomrule
		\end{tabular}
	}
	\label{tab:ablation_ksize}
\end{minipage}
\end{table}

\subsubsection{Single offline Teacher vs.\ Mutual online Teacher.} Table~\ref{tab:ablation_setting} shows the comparison between our proposed two variants of teacher-student learning. Both settings have greatly surpassed the performance of the base network (ID A3), showing the effectiveness of our proposed label densification strategy. We note that the mutual online teacher setting is $2.9\%$ higher than the single offline teacher setting. The reason could be that performance is bounded by the fixed teacher model while the mutual online teacher setting could improve each other over the training process.

\subsubsection{Effects of kernel size for label mask dilation.}
Table~\ref{tab:ablation_ksize} summarizes the results of different kernel size for label mask dilation. When increasing the kernel size, the performance rises first but then drops, with kernel size 7 being the best, which demonstrates the necessity of restricting pseudo-labels in a meaningful local neighborhood.

\section{Conclusion}

In this work, we propose a novel teacher-student learning paradigm in order to address the challenge of label sparsity for semantic correspondence task. In our teacher-student paradigm, we generate dense pseudo-labels by the teacher networks which are trained with sparse annotations. To improve quality of pseudo-labels, we develop two novel techniques to denoise pseudo-labels. Specifically, we first dilate the sparse label masks derived from the sparse keypoint annotations to suppress background pseudo-labels. A dynamic label selection strategy is then introduced to further filter noisy labels. We investigate two variants of the proposed learning paradigm, a single offline teacher setting, and a mutual online teacher setting. Our method achieves state-of-the-art performances on three standard datasets. The effectiveness of our method provides new insight into the problem, and is one step closer towards a more realistic application of semantic correspondence.

\clearpage

\bibliographystyle{splncs04}
\bibliography{2609}

\begin{thebibliography}{10}
\providecommand{\url}[1]{\texttt{#1}}
\providecommand{\urlprefix}{URL }
\providecommand{\doi}[1]{https://doi.org/#1}

\bibitem{blum1998combining}
Blum, A., Mitchell, T.: Combining labeled and unlabeled data with co-training.
  In: Proceedings of the Annual Conference on Learning Theory(COLT) (1998)

\bibitem{bromley1993signature}
Bromley, J., Guyon, I., LeCun, Y., S{\"a}ckinger, E., Shah, R.: Signature
  verification using a" siamese" time delay neural network. Advances in neural
  information processing systems  \textbf{6} (1993)

\bibitem{chauhan2013moving}
Chauhan, A.K., Krishan, P.: Moving object tracking using gaussian mixture model
  and optical flow. International Journal of Advanced Research in Computer
  Science and Software Engineering  \textbf{3}(4) (2013)

\bibitem{chen2015net2net}
Chen, T., Goodfellow, I., Shlens, J.: Net2net: Accelerating learning via
  knowledge transfer. In: Proceedings of the International Conference on
  Learning Representations(ICLR) (2015)

\bibitem{chen2020show}
Chen, Y.C., Lin, Y.Y., Yang, M.H., Huang, J.B.: Show, match and segment: Joint
  weakly supervised learning of semantic matching and object co-segmentation.
  IEEE Transactions on Pattern Analysis and Machine Intelligence  (2020)

\bibitem{cho2021cats}
Cho, S., Hong, S., Jeon, S., Lee, Y., Sohn, K., Kim, S.: Cats: Cost aggregation
  transformers for visual correspondence. In: Advances in Neural Information
  Processing Systems(NeurIPS) (2021)

\bibitem{dale2009image}
Dale, K., Johnson, M.K., Sunkavalli, K., Matusik, W., Pfister, H.: Image
  restoration using online photo collections. In: Proceedings of the
  International Conference on Computer Vision(ICCV) (2009)

\bibitem{goldstein2012video}
Goldstein, A., Fattal, R.: Video stabilization using epipolar geometry. ACM
  Transactions on Graphics (TOG)  \textbf{31}(5),  1--10 (2012)

\bibitem{ham2016proposal}
Ham, B., Cho, M., Schmid, C., Ponce, J.: Proposal flow. In: Proceedings of the
  IEEE Conference on Computer Vision and Pattern Recognition(CVPR) (2016)

\bibitem{ham2018proposal}
Ham, B., Cho, M., Schmid, C., Ponce, J.: Proposal flow: Semantic
  correspondences from object proposals. IEEE Transactions on Pattern Analysis
  and Machine Intelligence  (2018)

\bibitem{han2018co}
Han, B., Yao, Q., Yu, X., Niu, G., Xu, M., Hu, W., Tsang, I., Sugiyama, M.:
  Co-teaching: Robust training of deep neural networks with extremely noisy
  labels. In: Advances in Neural Information Processing Systems(NeurIPS) (2018)

\bibitem{khan2017}
Han, K., Rezende, R.S., Ham, B., Wong, K.Y.K., Cho, M., Schmid, C., Ponce, J.:
  Scnet: Learning semantic correspondence. In: Proceedings of the International
  Conference on Computer Vision(ICCV) (2017)

\bibitem{he2022asm}
He, B., Yang, X., Kang, L., Cheng, Z., Zhou, X., Shrivastava, A.: Asm-loc:
  Action-aware segment modeling for weakly-supervised temporal action
  localization. In: Proceedings of the IEEE Conference on Computer Vision and
  Pattern Recognition(CVPR) (2022)

\bibitem{he2020gta}
He, B., Yang, X., Wu, Z., Chen, H., Lim, S.N., Shrivastava, A.: Gta: Global
  temporal attention for video action understanding. Proceedings of the British
  Machine Vision Conference(BMVC)  (2020)

\bibitem{he2016deep}
He, K., Zhang, X., Ren, S., Sun, J.: Deep residual learning for image
  recognition. In: Proceedings of the IEEE Conference on Computer Vision and
  Pattern Recognition(CVPR) (2016)

\bibitem{heo2019comprehensive}
Heo, B., Kim, J., Yun, S., Park, H., Kwak, N., Choi, J.Y.: A comprehensive
  overhaul of feature distillation. In: Proceedings of the IEEE Conference on
  Computer Vision and Pattern Recognition(CVPR) (2019)

\bibitem{hinton2015distilling}
Hinton, G., Vinyals, O., Dean, J., et~al.: Distilling the knowledge in a neural
  network. arXiv preprint arXiv:1503.02531  \textbf{2}(7) (2015)

\bibitem{hong2022cost}
Hong, S., Cho, S., Nam, J., Lin, S., Kim, S.: Cost aggregation with 4d
  convolutional swin transformer for few-shot segmentation. In: Proceedings of
  the European Conference on Computer Vision(ECCV) (2022)

\bibitem{hongsuck2018attentive}
Hongsuck~Seo, P., Lee, J., Jung, D., Han, B., Cho, M.: Attentive semantic
  alignment with offset-aware correlation kernels. In: Proceedings of the
  European Conference on Computer Vision(ECCV) (2018)

\bibitem{horn1981determining}
Horn, B.K., Schunck, B.G.: Determining optical flow. Artificial intelligence
  \textbf{17}(1-3),  185--203 (1981)

\bibitem{huang2020confidence}
Huang, S., Wang, Q., He, X.: Confidence-aware adversarial learning for
  self-supervised semantic matching. In: Chinese Conference on Pattern
  Recognition and Computer Vision (PRCV). Springer (2020)

\bibitem{huang2019dynamic}
Huang, S., Wang, Q., Zhang, S., Yan, S., He, X.: Dynamic context correspondence
  network for semantic alignment. In: Proceedings of the IEEE International
  Conference on Computer Vision(ICCV) (2019)

\bibitem{huang2019ccnet}
Huang, Z., Wang, X., Huang, L., Huang, C., Wei, Y., Liu, W.: Ccnet: Criss-cross
  attention for semantic segmentation. In: Proceedings of the IEEE
  International Conference on Computer Vision(ICCV) (2019)

\bibitem{jeon2018parn}
Jeon, S., Kim, S., Min, D., Sohn, K.: Parn: Pyramidal affine regression
  networks for dense semantic correspondence. In: Proceedings of the European
  Conference on Computer Vision(ECCV) (2018)

\bibitem{jeon2020guided}
Jeon, S., Min, D., Kim, S., Choe, J., Sohn, K.: Guided semantic flow. In:
  Proceedings of the European Conference on Computer Vision(ECCV) (2020)

\bibitem{jeon2019joint}
Jeon, S., Min, D., Kim, S., Sohn, K.: Joint learning of semantic alignment and
  object landmark detection. In: Proceedings of the IEEE International
  Conference on Computer Vision(ICCV) (2019)

\bibitem{kim2018recurrent}
Kim, S., Lin, S., JEON, S.R., Min, D., Sohn, K.: Recurrent transformer networks
  for semantic correspondence. In: Advances in Neural Information Processing
  Systems(NeurIPS) (2018)

\bibitem{kim2017fcss}
Kim, S., Min, D., Ham, B., Jeon, S., Lin, S., Sohn, K.: Fcss: Fully
  convolutional self-similarity for dense semantic correspondence. In:
  Proceedings of the IEEE Conference on Computer Vision and Pattern
  Recognition(CVPR) (2017)

\bibitem{kim2019semantic}
Kim, S., Min, D., Jeong, S., Kim, S., Jeon, S., Sohn, K.: Semantic attribute
  matching networks. In: Proceedings of the IEEE Conference on Computer Vision
  and Pattern Recognition(CVPR) (2019)

\bibitem{lan2021discobox}
Lan, S., Yu, Z., Choy, C., Radhakrishnan, S., Liu, G., Zhu, Y., Davis, L.S.,
  Anandkumar, A.: Discobox: Weakly supervised instance segmentation and
  semantic correspondence from box supervision. In: Proceedings of the IEEE
  Conference on Computer Vision and Pattern Recognition(CVPR) (2021)

\bibitem{lee2021patchmatch}
Lee, J.Y., DeGol, J., Fragoso, V., Sinha, S.N.: Patchmatch-based neighborhood
  consensus for semantic correspondence. In: Proceedings of the IEEE Conference
  on Computer Vision and Pattern Recognition(CVPR) (2021)

\bibitem{lee2019sfnet}
Lee, J., Kim, D., Ponce, J., Ham, B.: Sfnet: Learning object-aware semantic
  correspondence. In: Proceedings of the IEEE Conference on Computer Vision and
  Pattern Recognition(CVPR) (2019)

\bibitem{lee2020reference}
Lee, J., Kim, E., Lee, Y., Kim, D., Chang, J., Choo, J.: Reference-based sketch
  image colorization using augmented-self reference and dense semantic
  correspondence. In: Proceedings of the IEEE Conference on Computer Vision and
  Pattern Recognition(CVPR) (2020)

\bibitem{li2022rethinking}
Li, H., Wu, Z., Shrivastava, A., Davis, L.S.: Rethinking pseudo labels for
  semi-supervised object detection. In: Proceedings of the AAAI Conference on
  Artificial Intelligence(AAAI) (2022)

\bibitem{li2020correspondence}
Li, S., Han, K., Costain, T.W., Howard-Jenkins, H., Prisacariu, V.:
  Correspondence networks with adaptive neighbourhood consensus. In:
  Proceedings of the IEEE Conference on Computer Vision and Pattern
  Recognition(CVPR) (2020)

\bibitem{li2021probabilistic}
Li, X., Fan, D.P., Yang, F., Luo, A., Cheng, H., Liu, Z.: Probabilistic model
  distillation for semantic correspondence. In: Proceedings of the IEEE
  Conference on Computer Vision and Pattern Recognition(CVPR) (2021)

\bibitem{liu2011sift}
Liu, C., Yuen, J., Torralba, A.: Sift flow: Dense correspondence across scenes
  and its applications. IEEE Transactions on Pattern Analysis and Machine
  Intelligence  (2011)

\bibitem{liu2020semantic}
Liu, Y., Zhu, L., Yamada, M., Yang, Y.: Semantic correspondence as an optimal
  transport problem. In: Proceedings of the IEEE Conference on Computer Vision
  and Pattern Recognition(CVPR) (2020)

\bibitem{min2021convolutional}
Min, J., Cho, M.: Convolutional hough matching networks. In: Proceedings of the
  IEEE Conference on Computer Vision and Pattern Recognition(CVPR) (2021)

\bibitem{min2019hyperpixel}
Min, J., Lee, J., Ponce, J., Cho, M.: Hyperpixel flow: Semantic correspondence
  with multi-layer neural features. In: Proceedings of the IEEE International
  Conference on Computer Vision(ICCV) (2019)

\bibitem{min2019spair}
Min, J., Lee, J., Ponce, J., Cho, M.: Spair-71k: A large-scale benchmark for
  semantic correspondence. arXiv preprint arXiv:1908.10543  (2019)

\bibitem{min2020learning}
Min, J., Lee, J., Ponce, J., Cho, M.: Learning to compose hypercolumns for
  visual correspondence. In: Proceedings of the European Conference on Computer
  Vision(ECCV) (2020)

\bibitem{paszke2017automatic}
Paszke, A., Gross, S., Chintala, S., Chanan, G., Yang, E., DeVito, Z., Lin, Z.,
  Desmaison, A., Antiga, L., Lerer, A.: Automatic differentiation in pytorch
  (2017)

\bibitem{Rocco2017}
Rocco, I., Arandjelovi{\'c}, R., Sivic, J.: Convolutional neural network
  architecture for geometric matching. In: Proceedings of the IEEE Conference
  on Computer Vision and Pattern Recognition(CVPR) (2017)

\bibitem{Rocco2018}
Rocco, I., Arandjelovi{\'c}, R., Sivic, J.: End-to-end weakly-supervised
  semantic alignment. In: Proceedings of the IEEE Conference on Computer Vision
  and Pattern Recognition(CVPR) (2018)

\bibitem{Rocco18b}
Rocco, I., Cimpoi, M., Arandjelovi\'{c}, R., Torii, A., Pajdla, T., Sivic, J.:
  Neighbourhood consensus networks. In: Advances in Neural Information
  Processing Systems(NeurIPS) (2018)

\bibitem{scharstein2002taxonomy}
Scharstein, D., Szeliski, R.: A taxonomy and evaluation of dense two-frame
  stereo correspondence algorithms. International Journal of Computer Vision
  \textbf{47}(1-3),  7--42 (2002)

\bibitem{sohn2020fixmatch}
Sohn, K., Berthelot, D., Carlini, N., Zhang, Z., Zhang, H., Raffel, C.A.,
  Cubuk, E.D., Kurakin, A., Li, C.L.: Fixmatch: Simplifying semi-supervised
  learning with consistency and confidence. In: Advances in Neural Information
  Processing Systems(NeurIPS) (2020)

\bibitem{taniai2016joint}
Taniai, T., Sinha, S.N., Sato, Y.: Joint recovery of dense correspondence and
  cosegmentation in two images. In: Proceedings of the IEEE Conference on
  Computer Vision and Pattern Recognition(CVPR) (2016)

\bibitem{tarvainen2017mean}
Tarvainen, A., Valpola, H.: Mean teachers are better role models:
  Weight-averaged consistency targets improve semi-supervised deep learning
  results. In: Advances in Neural Information Processing Systems(NeurIPS)
  (2017)

\bibitem{tola2010daisy}
Tola, E., Lepetit, V., Fua, P.: Daisy: An efficient dense descriptor applied to
  wide-baseline stereo. IEEE Transactions on Pattern Analysis and Machine
  Intelligence  (2010)

\bibitem{truong2020glu}
Truong, P., Danelljan, M., Timofte, R.: Glu-net: Global-local universal network
  for dense flow and correspondences. In: Proceedings of the IEEE Conference on
  Computer Vision and Pattern Recognition(CVPR) (2020)

\bibitem{truong2022probabilistic}
Truong, P., Danelljan, M., Yu, F., Van~Gool, L.: Probabilistic warp consistency
  for weakly-supervised semantic correspondences. In: Proceedings of the IEEE
  Conference on Computer Vision and Pattern Recognition(CVPR) (2022)

\bibitem{xie2020self}
Xie, Q., Luong, M.T., Hovy, E., Le, Q.V.: Self-training with noisy student
  improves imagenet classification. In: Proceedings of the IEEE Conference on
  Computer Vision and Pattern Recognition(CVPR) (2020)

\bibitem{yang2021deep}
Yang, L., Wang, Y., Gao, M., Shrivastava, A., Weinberger, K.Q., Chao, W.L.,
  Lim, S.N.: Deep co-training with task decomposition for semi-supervised
  domain adaptation. In: Proceedings of the International Conference on
  Computer Vision(ICCV) (2021)

\bibitem{yang2013articulated}
Yang, Y., Ramanan, D.: Articulated human detection with flexible mixtures of
  parts. IEEE Transactions on Pattern Analysis and Machine Intelligence  (2013)

\bibitem{yim2017gift}
Yim, J., Joo, D., Bae, J., Kim, J.: A gift from knowledge distillation: Fast
  optimization, network minimization and transfer learning. In: Proceedings of
  the IEEE Conference on Computer Vision and Pattern Recognition(CVPR) (2017)

\bibitem{yue2018compact}
Yue, K., Sun, M., Yuan, Y., Zhou, F., Ding, E., Xu, F.: Compact generalized
  non-local network. In: Advances in Neural Information Processing
  Systems(NeurIPS) (2018)

\bibitem{zhang2019latentgnn}
Zhang, S., He, X., Yan, S.: Latentgnn: Learning efficient non-local relations
  for visual recognition. In: Proceedings of the International Conference on
  Machine Learning(ICML) (2019)

\bibitem{zhang2018deep}
Zhang, Y., Xiang, T., Hospedales, T.M., Lu, H.: Deep mutual learning. In:
  Proceedings of the IEEE Conference on Computer Vision and Pattern
  Recognition(CVPR) (2018)

\bibitem{zhao2021multi}
Zhao, D., Song, Z., Ji, Z., Zhao, G., Ge, W., Yu, Y.: Multi-scale matching
  networks for semantic correspondence. In: Proceedings of the IEEE
  International Conference on Computer Vision(ICCV) (2021)

\bibitem{zhou2016learning}
Zhou, B., Khosla, A., Lapedriza, A., Oliva, A., Torralba, A.: Learning deep
  features for discriminative localization. In: Proceedings of the IEEE
  Conference on Computer Vision and Pattern Recognition(CVPR) (2016)

\end{thebibliography}

\end{document}